\pdfoutput=1

\documentclass[11pt]{article}

\usepackage[dvipsnames]{xcolor}
\usepackage{acl}

\usepackage{times}
\usepackage{latexsym}
\usepackage{multirow}
\usepackage{url}
\usepackage{amsmath}
\usepackage{booktabs}
\usepackage{graphicx}
\usepackage{enumitem}
\usepackage{amsfonts}

\usepackage{xurl}

\usepackage[T1]{fontenc}

\usepackage[utf8]{inputenc}

\usepackage{microtype}

\title{Multi-Granularity Semantic Aware Graph Model for Reducing Position Bias in Emotion-Cause Pair Extraction}

\author{Yinan Bao$^{1, 2}$, Qianwen Ma$^{1, 2}$, Lingwei Wei$^{1, 2}$, Wei Zhou$^{1,}$\thanks{\quad Corresponding author.} \and Songlin Hu$^{1, 2}$ \\
        $^{1}$Institute of Information Engineering, Chinese Academy of Sciences \\ $^{2}$School of Cyber Security, University of Chinese Academy of Sciences \\
        \texttt{\{baoyinan, maqianwen, weilingwei, zhouwei, husonglin\}@iie.ac.cn} \\}

\begin{document}
\maketitle


\begin{abstract}



The Emotion-Cause Pair Extraction (ECPE) task aims to extract emotions and causes as pairs from documents. We observe that the relative distance distribution of emotions and causes is extremely imbalanced in the typical ECPE dataset. Existing methods have set a fixed size window to capture relations between neighboring clauses. However, they neglect the effective semantic connections between distant clauses, leading to poor generalization ability towards position-insensitive data. To alleviate the problem, we propose a novel \textbf{M}ulti-\textbf{G}ranularity \textbf{S}emantic \textbf{A}ware \textbf{G}raph model (MGSAG) to incorporate fine-grained and coarse-grained semantic features jointly, without regard to distance limitation. In particular, we first explore semantic dependencies between clauses and keywords extracted from the document that convey fine-grained semantic features, obtaining keywords enhanced clause representations. Besides, a clause graph is also established to model coarse-grained semantic relations between clauses. Experimental results indicate that MGSAG surpasses the existing state-of-the-art ECPE models. Especially, MGSAG outperforms other models significantly in the condition of position-insensitive data.
\end{abstract}

\section{Introduction}

\begin{figure}[!htbp]
	\centering
	\includegraphics[scale=0.30]{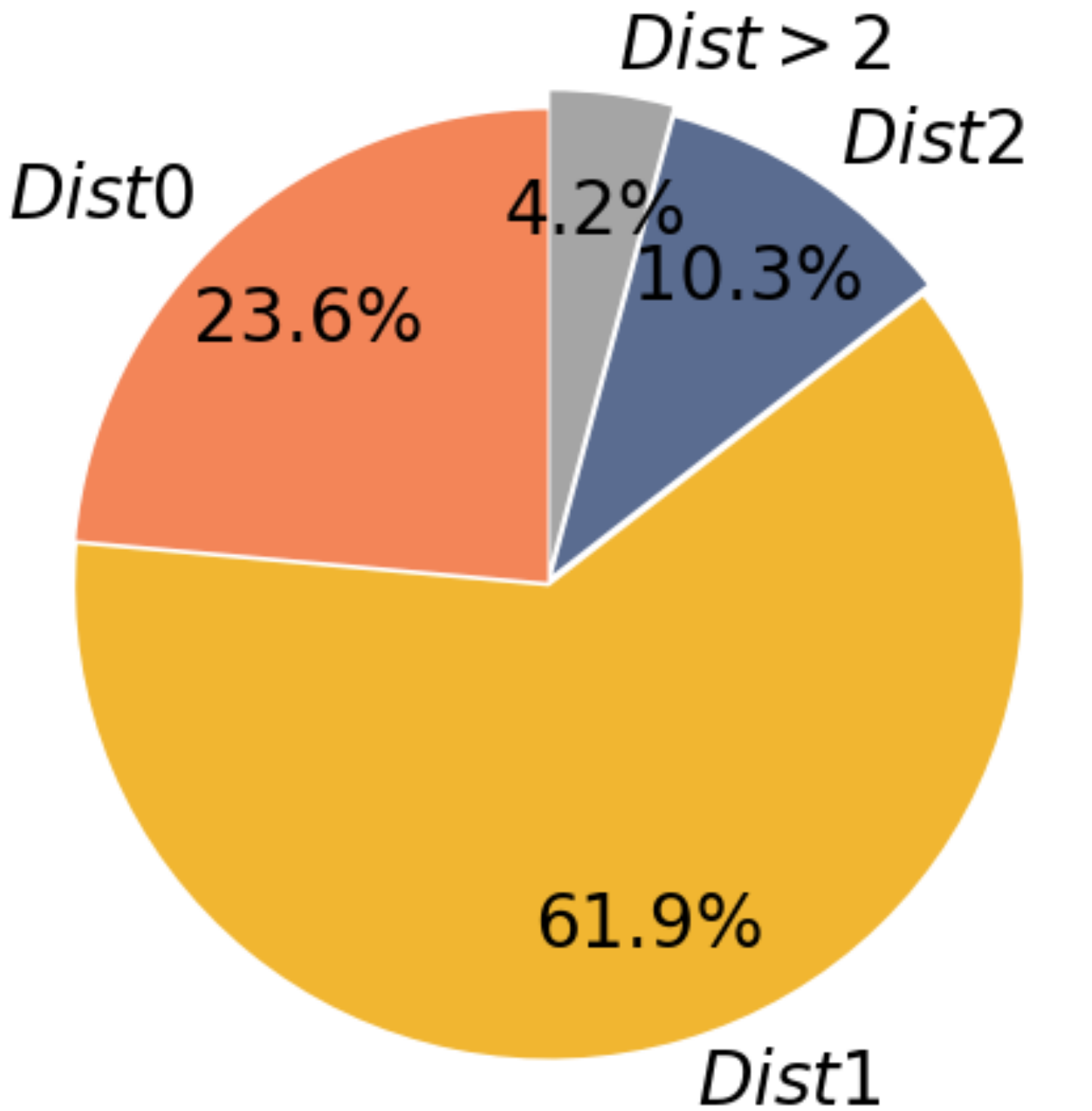}
	\caption{The distribution of the relative distance of an emotion clause and a cause clause that comprise a pair in the ECPE dataset \cite{ECPE:Xia19ACL}. $Dist0$, $Dist1$, and $Dist2$ mean the relative distances between the two clauses are 0, 1, and 2 respectively. $Dist>2$ means the relative distances are larger than 2.}
	\label{fig:pos_bias}
	\vspace{-1\baselineskip}
\end{figure}

Emotion Cause Analysis (ECA) has attracted increasing research interest in recent years \cite{RANKCP:Wei20ACL,DQAN:Sun21IJCNN,E2E_CExt:Singh21WASSA,MAM:Yu21IEEE}, because of the great potential of applying in consumer review mining, public opinion monitoring, and online empathetic chatbot building. Its goal is to detect causes or stimuli for a certain emotion expressed in text. 

Emotion Cause Pair Extraction (ECPE) ~\cite{ECPE:Xia19ACL} is a new task related to ECA, which is concerned with causal relationships between emotions and causes. It's a much more challenging task. Because we need a comprehensive understanding of document content and structure to perform emotion-cause co-extraction and discriminate emotion-cause clause pairs from negative ones ~\cite{RANKCP:Wei20ACL}. As shown in the following example, an emotion clause $c_7$ and a cause clause $c_2$ construct an emotion-cause pair $(c_7, c_2)$ which is needed to be extracted by an ECPE model.

\noindent\textbf{Example.} \textit{When the driver was about to start the bus to leave the station} ($c_1$), \underline{\textit{an old lady ran to the front of the bus with a fast}} \underline{\textit{speed and sat down on the ground}} ($c_2$). \textit{Passengers standing in the front of the bus can see this scene clearly} ($c_3$). \textit{Seeing this scene} ($c_4)$, \textit{the passengers in the car immediately became restless} ($c_5$), \textit{and had a heated debate} ($c_6$). \underline{\textit{Some of the passengers were \textbf{angry}}} ($c_7$), \textit{ and told the driver he shouldn't be meddlesome} ($c_8$).

In general, the number of candidate emotion-cause pairs is the square of the number of clauses in a document. However, most documents contain only one emotion-cause pair. Due to the problem of the tremendous search space, most existing methods have fully exploited relative position features to decrease the number of candidate pairs. For instance, ECPE-MLL \cite{ECPE-MLL:Ding20EMNLP} and SLSN \cite{SLSN:Cheng20COLING} set a fixed size window around a certain clause, and the central clause and other clauses inside the window comprise candidate pairs. However, models heavily relying on the relative position features ignore the distant semantic cues, resulting in poor generalization ability towards position-insensitive data in which the cause clause is not in proximity to the emotion clause.


According to Figure~\ref{fig:pos_bias}, we can observe that there is a position bias problem in ECPE. For the most 85\% emotion-cause pairs, the relative distances between its emotion clauses and corresponding cause clauses are less than 2. It means that most cause clauses either appear immediately preceding/following their corresponding emotion clauses or are the emotion clauses themselves. Existing methods mainly focus on the position-sensitive data (majority) and neglect the position-insensitive data (minority). How to improve the performance on the two parts of data instead of only focusing on one of them, has become an intractable challenge.






Some proposed methods~\cite{ECPE:Xia19ACL,IE-CNN:Chen20COLING} without relative position information seem to be position-insensitive, but overlook the effective semantic connections between distant clauses which convey causal cues. Thus, they can not alleviate the position bias problem. 
To alleviate this problem, we propose a multi-granularity semantic aware graph model (MGSAG). We assume that fine-grained semantic features conveyed by global keywords in a document are conducive to exploring causal cues, especially cues implied in distant clauses. Besides, coarse-grained semantics between clauses is also important to find causal relations implied in the context. From the two perspectives, we realize multi-granularity semantic enhanced clause relationships modeling based on two graphs: clause-keyword bipartite graph and fully connected clause graph, utilize fine-grained and coarse-grained semantic features jointly. Experimental results show that MGSAG outperforms all of the state-of-the-art baselines. Especially, it achieves a significant improvement on position-insensitive test data. In summary, our contributions are three-fold:

\begin{itemize}





\item  To alleviate the position bias problem in ECPE, we propose MGSAG to achieve fine-grained and coarse-grained semantic enhanced clause representation learning.

\item To value model performance on emotion-cause clause pairs consisting of distant clauses, we split the original test set into two parts according to the relative distances of emotion clauses and cause clauses, and evaluate models on them.

\item Experimental results prove that our model achieves remarkable improvement over best-performing approaches on the original test set. Especially, it outperforms other methods in the condition of position-insensitive data.


\end{itemize}

\section{Related Work}


According to whether the relative position information is used explicitly or not, existing ECPE works can be divided into two categories: position-sensitive approaches and position-insensitive approaches.


\noindent{\textbf{Position-Sensitive Approaches.}} Most methods~\cite{ECPE-2D:Ding20ACL,SLSN:Cheng20COLING,ECPE-MLL:Ding20EMNLP} have set a fixed size window to reduce the number of candidate pairs according to the inherent position bias in the dataset, because of the sparsity of true emotion-cause pairs compared with candidate emotion-cause pairs. Besides, \citet{PairGCN:ChenZ20COLING} leveraged the relative position information explicitly in the process of pair representation learning. The ECPE-MLL model proposed by \citet{ECPE-MLL:Ding20EMNLP} is the state-of-the-art method in the ECPE task. An over-reliance on relative position information makes these methods have poor generalization ability towards position-insensitive data.

\noindent{\textbf{Position-Insensitive Approaches.}} Some sequence-based methods without relative position information \cite{ECPE:Xia19ACL,IE-CNN:Chen20COLING,Trans:Fan20ACL} seem to be position-insensitive. \citet{ECPE:Xia19ACL} proposed a RNN-based framework and generate candidate pairs by applying the Cartesian product. \citet{IE-CNN:Chen20COLING} reformulated the ECPE task as a unified sequence labeling problem. \citet{Trans:Fan20ACL} modeled the extraction of emotion-cause pairs as performing a sequence of transitions and actions. However, these methods have shown poor performance on position-insensitive data due to the neglect of effective semantic connections between distant clauses.

Different from the above methods, our model incorporates fine-grained and coarse-grained semantic features jointly, which can alleviate the position bias problem well.
\begin{figure*}[!htbp]
	\centering
	\includegraphics[scale=0.15]{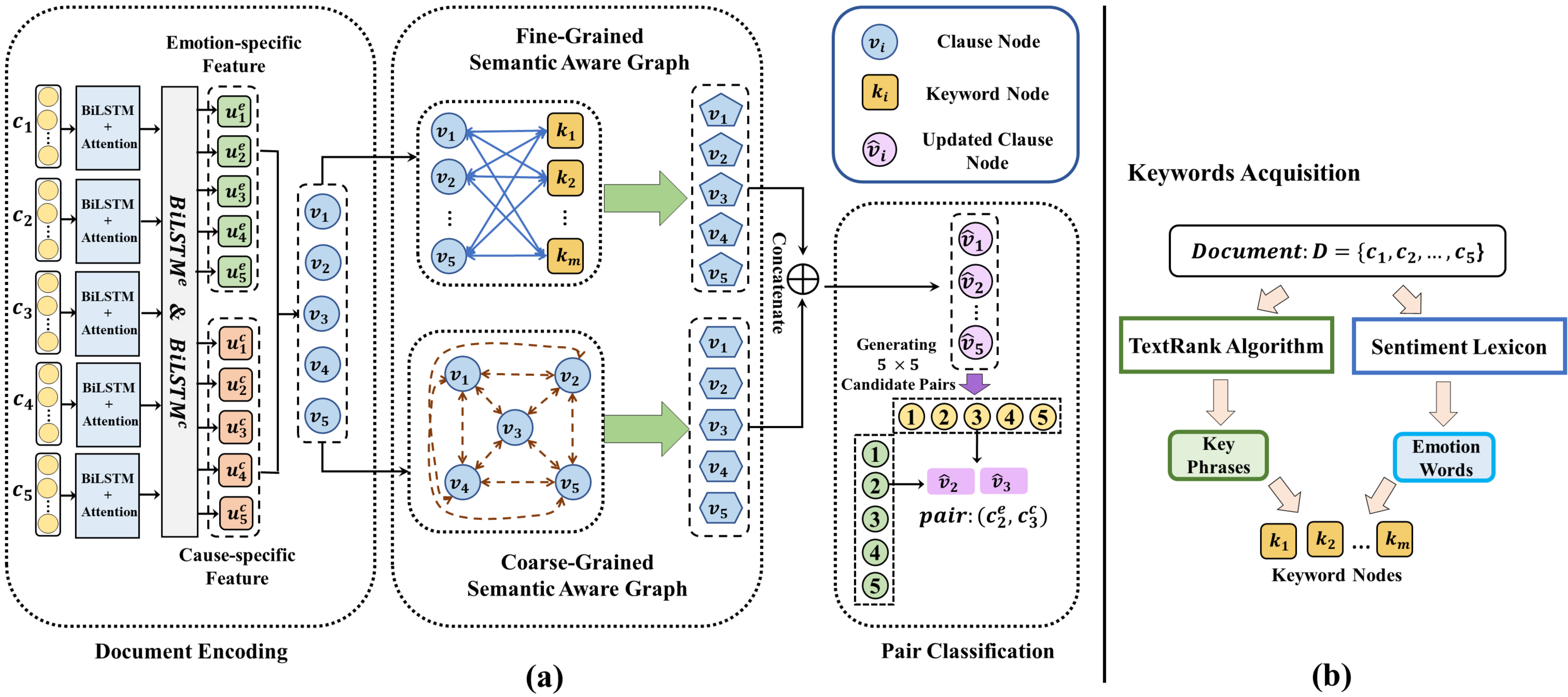}
	\vspace{-0.5\baselineskip}
	\caption{(a) shows an overview of MGSAG. (b) shows the process of keywords acquisition.}
	\label{fig:model}
\end{figure*}

\section{Problem Formulation}

Given a document $D = {\{c_1, c_2, ..., c_{|D|}}\}$ where $|D|$ is the number of clauses, the clauses are formed into $|D|\times |D|$ candidate emotion-cause pairs using Cartesian product: $P = {\{..., (c^e_i, c^c_j), ...}\}$, where $c^e_i$ is clause $c_i$ serving as a candidate emotion clause, $c^c_j$ is clause $c_j$ serving as a candidate cause clause. The ECPE task is to assign a binary label to each candidate pair $(c^e_i, c^c_j)$, where ``1'' means that clause $c_i$ is an emotion clause and clause $c_j$ provides the cause of it, otherwise ``0''.

\section{Methodology}

We propose a multi-granularity semantic aware graph model to alleviate the position bias problem in ECPE. More concretely, we obtain fine-grained semantic aware clause representations based on a clause-keyword bipartite graph. Simultaneously, coarse-grained semantic aware clause representations are generated based on a fully connected clause graph. As shown in Figure~\ref{fig:model}, the model consists of four components: 1) document encoding, 2) fine-grained semantic aware graph (FGSAG), 3) coarse-grained semantic aware graph (CGSAG), 4) pair classification.

\subsection{Document Encoding}

Given a document $D = {\{c_1, c_2, ..., c_{|D|}}\}$ consisted of $|D|$ clauses, we adopt a hierarchical recurrent neural network to encode context information and generate emotion-specific and cause-specific clause representations for each clause in the document.


\noindent{\textbf{Word-Level Encoder.}} For each clause $c_i = {\{w^i_1, w^i_2, ..., w^i_{|c_i|}}\}$, we first adopt a word-level BiLSTM network to encode the context by passing words' information along the clauses forwards and backwards, and then obtain the clause's hidden state sequence $(h^i_1, h^i_2, ..., h^i_{|c_i|})$. An attention layer is adopted to combine them and return a state vector $\mathbf{h}_i = \sum^{|c_i|}_{j=1} \alpha_j h^i_j$ for clause $c_i$, where $\alpha_j = \mathbf{softmax}(\mathbf{W}_a h^i_j)$ is the attention weight of the $j$-th word in clause $c_i$, $\mathbf{W}_a$ is a trainable weight matrix for attention score calculation.


\noindent{\textbf{Clause-Level Encoder.}} In order to extract the emotion features and the cause features respectively, the clause-level encoder consists of two BiLSTM networks. The document $D$'s clause state sequence $(\mathbf{h}_1, \mathbf{h}_2, ..., \mathbf{h}_{|D|})$ is fed into two clause-level BiLSTM networks to produce emotion-specific and cause-specific clause representations, respectively:\begin{equation} \label{equ:clause_level_encoder}
\begin{split}
& \mathbf{u}^e_i = \mathbf{BiLSTM}^e(\mathbf{h}_i)  \,, \\
& \mathbf{u}^c_i = \mathbf{BiLSTM}^c(\mathbf{h}_i) \,, \\
\end{split}
\end{equation}
where $\mathbf{BiLSTM}^e$ and $\mathbf{BiLSTM}^c$ generate the emotion-specific and cause-specific clause representation $\mathbf{u}^e_i,\mathbf{u}^c_i \in \mathbb{R}^{2d_h\times 1}$ of clause $c_i$, respectively. $d_h$ means the number of hidden units in BiLSTM.


Afterwards, we use a gate mechanism to fuse the emotion feature $\mathbf{u}^e_i$ and the cause feature $\mathbf{u}^c_i$ to obtain clause representation $\mathbf{v}_i\in \mathbb{R}^{2d_h\times 1}$:
\begin{equation} \label{equ:gate_fuse}
\begin{split}
& \mathbf{g}_i = \mathbf{\sigma}(\mathbf{W}_g \mathbf{u}^e_i + \mathbf{b}_g) \,, \\
& \mathbf{v}_i = \mathbf{g}_i \mathbf{u}^c_i + (1 - \mathbf{g}_i) \mathbf{u}^e_i  \,, \\
\end{split}
\end{equation}
where $\mathbf{W}_g \in R^{1\times 2d_h}$ and $\mathbf{b}_g$ are parameters; $\sigma$ is the sigmoid function.

In the training process, we leverage the emotion labels and cause labels as auxiliary supervision signals to facilitate the clause representation learning in the clause-level encoder:
\begin{equation} \label{equ:auxiliary_supervision}
\begin{split}
& \mathbf{\hat{y}}^e_i = \mathbf{softmax}(\mathbf{W}_e \mathbf{u}^e_i + \mathbf{b}_e)  \,, \\
& \mathbf{\hat{y}}^c_i = \mathbf{softmax}(\mathbf{W}_c \mathbf{u}^c_i + \mathbf{b}_c) \,, \\
\end{split}
\end{equation}
where $\mathbf{W}_e, \mathbf{W}_c \in \mathbb{R}^{1\times 2d_h}$ are trainable parameters and $\mathbf{b}_e, \mathbf{b}_c$ are bias terms.

\subsection{Fine-Grained Semantic Aware Graph}


To obtain fine-grained semantic enhanced clause representations, we leverage external knowledge to extract keywords in the document first. Then, we build a clause-keyword bipartite graph to model the relations between clauses. In this way, the keywords which convey fine-grained semantic features can help highlight the potential causal features contained in the clause representations.

\begin{figure}[!tbp]
	\centering
	\includegraphics[scale=0.20]{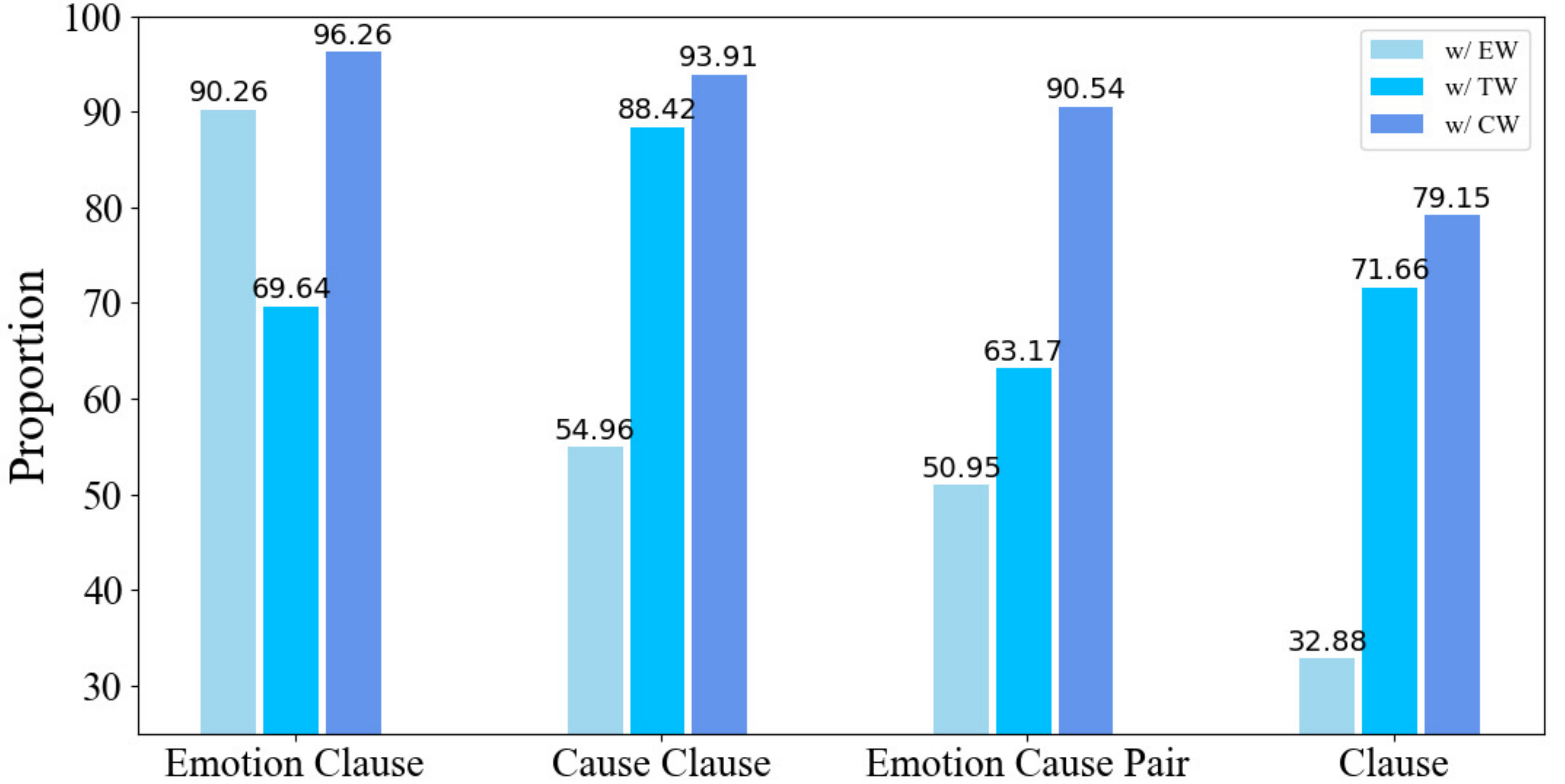}
	\caption{The influence of the two types of keywords from an intuitive aspect. It shows the proportion of emotion clauses, cause clauses, emotion-cause pairs, and clauses that are covered by the extracted key phrases or emotion words or both of them. ``w/ EW'', ``w/ TW'', and ``w/ CW'' means using emotion words, key phrases obtained by TextRank or both of them, respectively.}
	\label{fig:keywords}
	\vspace{-1\baselineskip}
\end{figure}

\noindent{\textbf{Keywords Acquisition.}} We use the TextRank algorithm \cite{TextRank:Mihalcea04EMNLP} to extract key phrases and a sentiment lexicon \cite{SentiDict:Xu08}\footnote{We download the sentiment lexicon from this link: \url{https://github.com/ZaneMuir/DLUT-Emotionontology}.} to obtain emotion words in a document. We take the union of the two sets as the final keyword set.

To measure the influence of the two types of keywords from an intuitive view, we count the proportions of emotion clauses, cause clauses, emotion-cause pairs, and clauses that are covered by the emotion words or key phrases or both of them. Noted that if emotion clause and cause clause that comprise a pair both contain any keyword, we think that the pair is covered by the keywords.  
From Figure~\ref{fig:keywords} we observe that if we use the key phrases extracted by TextRank alone, only about 69\% of emotion clauses can be found; if we use the emotion words alone, only about 54\% of cause clauses can be identified. With the use of emotion words or key phrases, only about 50\% or 63\% of emotion-cause pairs can be figured out. Consequently, we take the union of the two sets as the final keyword set. However, given the complete keyword set, clauses that contain keywords account for a large proportion (79\%), which means that the imported keywords may introduce noise as well. To this end, it's necessary to measure the importance of different keywords when modeling the interaction between clauses and keywords. 


\noindent{\textbf{Clause-Keyword Bipartite Graph Construction.}} Given a document $D$, we denote the clause-keyword bipartite graph as $\mathcal{G}_b = (\mathcal{V}, \mathcal{E}_b)$, where $\mathcal{V} = \mathcal{V}_c \cup \mathcal{V}_k$ represents a node set composing of clause nodes and keyword nodes and $\mathcal{E}_b$ denotes edges between nodes. $\mathcal{V}_k = {\{k_1, k_2, ..., k_m}\}$ and $\mathcal{V}_c = {\{c_1, c_2, ..., c_{|D|}}\}$ mean there are $m$ keywords and $|D|$ clauses in the document $D$. We establish edges between each node in $\mathcal{V}_c$ and each node in $\mathcal{V}_k$, which means every element $e_{ij}$ in $\mathcal{E}_b\in \mathbb{R}^{|D|\times m}$ is 1. It is because the average length of clauses is too short, many keywords only appear once in one clause. Thus, an adjacency matrix based on keyword-clause co-occurrence is extremely sparse.



For keywords in $\mathcal{V}_k$, their feature vectors are initialized by the word embedding vectors released by \citet{ECPE:Xia19ACL}. As for clause nodes $c_i\in \mathcal{V}_c$, they are initialized with the corresponding context-aware clause representation $\mathbf{v}_i$ generated from the clause-level encoder. We denote the feature matrices of keyword and clause nodes as $\mathbf{X}_k ={\{\mathbf{k}_1, ..., \mathbf{k}_m}\} \in \mathbb{R}^{m\times d_w}$ and $\mathbf{X}_c ={\{\mathbf{v}_1, ..., \mathbf{v}_{|D|}}\} \in \mathbb{R}^{|D|\times 2d_h}$ respectively, where $d_w$ is the dimension of the word embedding and is equal to $2d_h$ in our setting.


\noindent{\textbf{Attention Guided Clause Representations Update.}} We propose a graph attention module to model the semantic interaction between clauses and keywords, aiming to utilize the fine-grained semantic features implied in keywords to facilitate clause representation learning.

Intuitively, the clause-keyword bipartite graph realizes fine-grained semantic connections between distant clauses, which is helpful to extract emotion-cause pairs composed of distant clauses. Nevertheless, for a specific clause, the importance of various keywords is different. Therefore, we use the graph attention mechanism \cite{GAN:Velickovic17} to measure the document-level keyword preference-degree of each clause, where the attention weight is computed as the edge weight between the clause node $c_i$ and the keyword node $k_j$ in a document:
\begin{equation} \label{equ:bipartite_att}
\begin{split}
& \alpha_{ij} = \frac{\mathbf{exp}(w^\top [ \mathbf{W}_1 \mathbf{v}_i; \mathbf{W}_2 \mathbf{k}_j ])}{\sum^{|D|}_{t=1} \mathbf{exp}(w^\top [ \mathbf{W}_1 \mathbf{v}_t; \mathbf{W}_2 \mathbf{k}_j ]) } \,, \\
\end{split}
\end{equation}
where $\mathbf{v}_i$ and $\mathbf{k}_j$ are features of clause $c_i$ and keyword $k_j$ respectively; $[\cdot; \cdot]$ is the concatenation operation; $\mathbf{W}_1, \mathbf{W}_2\in \mathbb{R}^{d_w\times d_w}$ and $w\in \mathbb{R}^{2d_w\times 1}$ are trainable parameters. 

Then, clause $c_i$ is encoded as the fine-grained semantic enhanced representation $\mathbf{v}^b_i$ as follows:
\begin{equation} \label{equ:bipartite_rep}
\begin{split}
& \mathbf{v}^b_i\! =\! \mathbf{tanh}(( \mathbf{v}_i + \sum^m_{j=1}( \alpha_{ij}(\sum^{|D|}_{t=1} \alpha_{tj} \mathbf{W}_3 \mathbf{v}_t ) ))\! +\! \mathbf{b}) \,, \\
\end{split}
\end{equation}
where $\sum^{|D|}_{t=1} \alpha_{tj} \mathbf{W}_3 \mathbf{v}_t$ means the representation of the keyword $\mathbf{k}_j$, and $\sum^m_{j=1}( \alpha_{ij}(\sum^{|D|}_{t=1} \alpha_{tj} \mathbf{W}_3 \mathbf{v}_t ) )$ is the weighted added of keyword representations for generating fine-grained semantic enhanced clause representation. $\mathbf{W}_3\in \mathbb{R}^{d_w\times d_w}$ is a trainable parameter and $\mathbf{b}$ is a bias term.



\subsection{Coarse-Grained Semantic Aware Graph}

Coarse-grained semantic relationships between clauses are useful for finding causal cues implied in the context. We establish a fully connected clause graph and leverage graph attention mechanism to model the coarse-grained semantic relationships between clauses.

Given a document $D$, we define the clause graph as $\mathcal{G}_c = (\mathcal{V}_c, \mathcal{E}_c)$, where $\mathcal{V}_c$ represents a node set and $\mathcal{E}_c$ denotes an edge set. Each node in the fully connected graph is a clause in $D$, and every two nodes have an edge. Self-loop edge is added to every node because a clause can be an emotion clause and a cause clause simultaneously. We use clause representation $\mathbf{v}_i$ generated from the clause-level encoder for node feature initialization. Based on the self-attention mechanism \cite{Self_Att:Vaswani17NIPS} which aggregated neighboring clauses' information, the graph attention network propagates information among clauses by stacking multiple graph attention layers. The representation of clause $c_i$ in the $t$-th layer is updated as follows:
\begin{equation} \label{equ:clause_gat}
\begin{split}
& \mathbf{v}^{(t)}_i = \mathbf{ReLU}(\sum_{j\in \mathcal{N}(i)}\alpha^{(t)}_{ij} \mathbf{W}^{(t)}_1\mathbf{v}^{(t-1)}_j + \mathbf{b}^{(t)} ) \,, \\
\end{split}
\end{equation}
where $\mathbf{W}^{(t)}_1\in \mathbb{R}^{d_w\times d_w}$ is a transform matrix and $\mathbf{b}^{(t)}$ is a bias term; $\mathcal{N}(i)$ represents the neighbouring clauses of $c_i$; $\mathbf{v}^{(0)}_i = \mathbf{v}_i$. The attention weight $\alpha^{(t)}_{ij}$ is learned as follows:
\begin{equation} \label{equ:clause_att}
\begin{split}
& e^{(t)}_{ij} = {w^{(t)}}^\top \mathbf{tanh}([ \mathbf{W}^{(t)}_2 \mathbf{v}^{(t-1)}_i; \mathbf{W}^{(t)}_3 \mathbf{v}^{(t-1)}_j ]) \,, \\
& \alpha^{(t)}_{ij} = \frac{\mathbf{exp}(\mathbf{LeakyReLU}(e^{(t)}_{ij}))}{\sum_{k\in \mathcal{N}(i)} \mathbf{exp}(\mathbf{LeakyReLU}(e^{(t)}_{ik})) } \,, \\
\end{split}
\end{equation}

We stack two graph attention layers and obtain $\mathbf{v}^c_i = \mathbf{v}^{(2)}_i$ as the updated representation for $c_i$.

\subsection{Pair Classification}

We concatenate the two types of clause representations and obtain $\mathbf{\hat{v}}_i = [\mathbf{v}^b_i; \mathbf{v}^c_i]$ as the final representation of clause $c_i$.

\noindent{\textbf{Emotion Cause Pair Extraction.}} For a candidate pair $(c^e_i, c^c_j)\in P$, we pass its representation $\mathbf{v}^p_{ij} = [\mathbf{\hat{v}}_i; \mathbf{\hat{v}}_j]$ to a fully-connected layer with softmax activation function to predict the label of it:
\begin{equation} \label{equ:ecp_extraction}
\begin{split}
& \hat{\mathbf{p}}_{ij} = \mathbf{softmax}( \mathbf{W}^\top_p \mathbf{v}^p_{ij} + \mathbf{b}_p ) \,, \\
\end{split}
\end{equation}
where $\mathbf{W}_p\in \mathbb{R}^{4d_w\times 2}$ and $\mathbf{b}_p\in \mathbb{R}^{2\times 1}$ are trainable parameters. We obtain the predicted label $\hat{\mathbf{EC}}_{ij}$ for the candidate pair $(c^e_i, c^c_j)$ according to the probability distribution $\hat{\mathbf{p}}_{ij}$.

During model training, we use two cross-entropy loss functions $\mathcal{L}_{emo}$ and $\mathcal{L}_{cau}$ to supervise the clause representation learning in the clause-level encoder and a cross-entropy loss function $\mathcal{L}_{pair}$ to supervise the final emotion-cause pair prediction. The loss function $\mathcal{L}$ is formulated as follows:
\begin{equation} \label{equ:loss_function}
\begin{split}
& \mathcal{L} = \mathcal{L}_{pair} + \mathcal{L}_{emo} + \mathcal{L}_{cau} \,. \\
\end{split}
\end{equation}

\noindent{\textbf{Emotion Extraction and Cause Extraction.}} Following \citet{PairGCN:ChenZ20COLING}, we implement emotion extraction and cause extraction based on the predictions of all candidate pairs. For emotion extraction, the predicted label $\hat{\mathbf{E}}_i$ for clause $c_i$ can be obtained as follows:
\begin{equation} \label{equ:emotion_extraction}
\begin{split}
& \hat{\mathbf{E}}_i = \left\{
\begin{array}{rcl}
1,  &   & {if\sum^{|D|}_{j=1}(\hat{\mathbf{EC}}_{ij}) > 0} \\
0,  &   & otherwise
\end{array}
\right. \,. \\
\end{split}
\end{equation}

For cause extraction, the predicted label $\hat{\mathbf{C}}_i$ for clause $c_i$ can be obtained similarly.


\section{Experiments}

We conduct a series of experiments to verify the effectiveness of MGSAG.

\subsection{Experimental Setup}

\subsubsection{Dataset and Evaluation Metrics}

We use the benchmark dataset released by \citet{ECPE:Xia19ACL} for experiments. This typical and widely used dataset is constructed based on an emotion cause extraction corpus \cite{ECE:Gui16EMNLP} that contains 1,945 Chinese documents from SINA city news\footnote{\url{http://news.sina.com.cn/society/}
}. To obtain statistically credible results, we adopt the same data split setting (10-fold cross-validation) used by \citet{ECPE:Xia19ACL}, repeat the experiments 10 times, and report the average results of precision (P), recall (R), and $F_1$-score ($F_1$) on the main task: emotion-cause pair extraction (\textbf{ECPE}), and two sub-tasks: emotion extraction (\textbf{EE}) and cause extraction (\textbf{CE}), following existing works \cite{ECPE:Xia19ACL,ECPE-MLL:Ding20EMNLP,ECPE-2D:Ding20ACL,IE-CNN:Chen20COLING,PairGCN:ChenZ20COLING,SLSN:Cheng20COLING}.



\subsubsection{Redistricting of Original Test Set}


As ECPE is a newly proposed task, there is only one typical and widely used dataset. Because of the inherent position bias in ECPE, how to improve the performance on both position-sensitive (majority) and position-insensitive data (minority), has become one of the challenges. Therefore, it is essential to measure the reliance of existing methods on the relative position information.

To this end, we split the original test set ($Test_{all}$) of each fold into two parts according to the relative distance between emotions and causes. The first part ($Test_{Bias}$) contains documents with only one pair and the relative distance between the two clauses is less than 2. The second part ($Test_{NoBias}$) is the complement of the first part, which means $Test_{all} = Test_{Bias} \cup Test_{NoBias}$ and $Test_{Bias} \cap Test_{NoBias} = \emptyset$. We conduct experiments on the original test set first, and then use $Test_{Bias}$ and $Test_{NoBias}$ to evaluate various methods respectively. To ensure fairness, we use the same model parameters which produce results on $Test_{all}$ to obtain the results on the two subsets: $Test_{Bias}$ and $Test_{NoBias}$. 


\subsubsection{Comparative Approaches}


We compare MGSAG with the following methods, which can be divided into two types: position-insensitive and position-sensitive methods.



\noindent\textbf{Position-insensitive Methods.} Following methods haven't utilized the relative position information explicitly. \textbf{Indep / Inter-CE / Inter-EC} \cite{ECPE:Xia19ACL}: these two-step approaches first extracted emotions and causes separately to form candidate emotion-cause pairs and then trained a classifier to recognize true pairs. \textbf{IE-CNN} \cite{IE-CNN:Chen20COLING} reformulated the ECPE task as a sequence labeling task and extracted pairs in an end-to-end fashion.

    


\noindent\textbf{Position-sensitive Methods.} Following methods take relative position information as a crucial feature to recognize pairs. \textbf{PairGCN} \cite{PairGCN:ChenZ20COLING} is a method highly dependent on position information when modeling relations between pairs. \textbf{ECPE-2D} \cite{ECPE-2D:Ding20ACL} extracted pairs through 2D representation, interaction, and prediction. The window-constrained 2D Transformer achieved the best performance. \textbf{SLSN-U} \cite{SLSN:Cheng20COLING} extracted pairs through a process of local search which was defined by the setting of the local context window. \textbf{RankCP} \cite{RANKCP:Wei20ACL} utilized kernel-based relative position embedding to enhance the clause representations obtained from inter-clause modeling module. \textbf{ECPE-MLL} \cite{ECPE-MLL:Ding20EMNLP} used a multi-label learning method inside each sliding window which was defined manually.

\subsubsection{Implementation Details}

To conduct a fair comparison with the baselines, we utilize the same word embeddings followed \citet{ECPE:Xia19ACL}. The dimension of word embedding is 200. The numbers of hidden units of BiLSTM in the word-level and clause-level encoder are set to 200 and 100, respectively. We stack two graph attention layers to build a graph attention network and add dropout \cite{Dropout:Srivastava14} with the rate of 0.1 for each layer to reduce over-fitting. During the training process, we use the Adam \cite{Adam:Kingma14ICLR} optimizer to update all parameters. We report the results of BERT \cite{BERT:Devlin19NAACL} in the appendix.




\begin{table*}[!htbp]\small
    \centering
    \renewcommand\arraystretch{1.0}
    \setlength{\tabcolsep}{5pt}{
    \begin{tabular}{c|l|ccc|ccc|ccc}
    \toprule
        \multirow{2}{*}{Category} & \multirow{2}{*}{Model} & \multicolumn{3}{|c|}{\textbf{Emotion Ext.}} & \multicolumn{3}{|c|}{\textbf{Cause Ext.}} & \multicolumn{3}{|c}{\textbf{EC Pair Ext.}} \\
        \cline{3-11}
            ~ & ~ & P & R & $F_1$ & P & R & $F_1$ & P & R & $F_1$ \\
         
         
         \midrule[0.7pt]
         \multirow{4}{*}{\centering \shortstack{Position-insensitive \\ Baselines}} & 
         Indep & 0.8375 & 0.8071 & 0.8210 & 0.6902 & 0.5673 & 0.6205 & 0.6832 & 0.5082 & 0.5818  \\
         ~ & Inter-CE & 0.8494 & 0.8122 & 0.8300  & 0.6809 & 0.5634 & 0.6151 & 0.6902 & 0.5135 & 0.5901  \\
         ~ & Inter-EC & 0.8364 & 0.8107 & 0.8230  & 0.7041 & 0.6083 & 0.6507 & 0.6721 & 0.5705 & 0.6128  \\
         ~ & IE-CNN & 0.8614 & 0.7811 & 0.8188  & 0.7348 & 0.5841 & 0.6496 & 0.7149 & 0.6279 & 0.6686  \\
        
        \midrule[0.3pt]
        
        \multirow{5}{*}{\shortstack{Position-sensitive \\ Baselines}} & PairGCN & 0.8587 & 0.7208 & 0.7829 & 0.7283 & 0.5953 & 0.6541 & 0.6999 & 0.5779 & 0.6321  \\
        ~ & ECPE-2D & 0.8512 & 0.8220 & 0.8358 & 0.7272 & 0.6298 & 0.6738 & 0.6960 & 0.6118 & 0.6496  \\
        ~ & SLSN-U & 0.8406 & 0.7980 &  0.8181 & 0.6992 & 0.6588 & 0.6778 & 0.6836 & 0.6291 &  0.6545 \\
        ~ & RankCP & 0.8703 & 0.8406 &  \textbf{0.8548} & 0.6927 & \textbf{0.6743} & 0.6824 & 0.6698 & \textbf{0.6546} &  0.6610 \\
        ~ & ECPE-MLL & 0.8582 & \textbf{0.8429} & 0.8500 & 0.7248 & 0.6702 & 0.6950 & 0.7090 & 0.6441 & 0.6740  \\
        
        \midrule[0.3pt]
        Our Model & \textbf{MGSAG} & \textbf{0.8721} & 0.7911 & 0.8287 & \textbf{0.7510} & 0.6713 & \textbf{0.7080} & \textbf{0.7243} & 0.6507 & \textbf{0.6846} \\
         
        
        
        
         

        
    \bottomrule
    \end{tabular}
    }
    \caption{Comparison of varying approaches on the original test set ($Test_{all}$).}
    \label{tab:All_Test_Results}
\end{table*}

\begin{table}[!htbp]\small
    \centering
    \renewcommand\arraystretch{1.0}
    \setlength{\tabcolsep}{15pt}{
    \begin{tabular}{l|c|c}
    \toprule
          Model & $Test_{Bias}$ & $Test_{NoBias}$ \\
        \midrule[0.7pt]
         
         
         Inter-EC & 0.6783  & 0.3318  \\
         IE-CNN & 0.7666 & 0.3484  \\
        
        \midrule[0.3pt]
        PairGCN & 0.7246 & 0.3355  \\
        ECPE-2D & 0.7590 & 0.3830  \\
        SLSN-U & 0.7456 & 0.3978  \\
        
        RankCP & 0.7467 & 0.3857 \\
        
        ECPE-MLL & 0.7673 & 0.3988  \\
        
        \midrule[0.3pt]
        \textbf{MGSAG} &  \textbf{0.7730} & \textbf{0.4301} \\
         

        
        
         

        
    \bottomrule
    \end{tabular}
    }
    \caption{$F_1$ results of varying approaches on $Test_{Bias}$ and $Test_{NoBias}$, focusing on EC Pair Ext.}
    \label{tab:Test1_Test2_Results}
\end{table}

\begin{table}[!htbp]\small
    \centering
    \renewcommand\arraystretch{1.0}
    \setlength{\tabcolsep}{3pt}{
    \begin{tabular}{l|c|c|c}
    \toprule
        Model & $Test_{Bias}$ & $Test_{NoBias}$ & $Test_{all}$ \\
         
         
        \midrule[0.7pt]
          
        w/o FGSAG & 0.7594  &  0.3894 & 0.6519 \\
        \midrule[0.3pt]
        
        w/o CGSAG & 0.7654  & 0.4027 & 0.6529 \\
        \midrule[0.3pt]
        
        w/o FGSAG+CGSAG & 0.7264 & 0.3269 & 0.6242 \\
         
        \midrule[0.7pt]

        \textbf{MGSAG} & \textbf{0.7730} &  \textbf{0.4301} & \textbf{0.6846} \\
        
    \bottomrule
    \end{tabular}
    }
    \caption{$F_1$ results of ablation study on $Test_{Bias}$, $Test_{NoBias}$, and $Test_{all}$, focusing on EC Pair Ext.}
    \label{tab:Ablation_Study}
    \vspace{-1\baselineskip}
\end{table}

\subsection{Experimental Results}

\subsubsection{Results on Original Test Set}

Table~\ref{tab:All_Test_Results} reports the comparative results on emotion cause pair extraction and two sub-tasks. We can observe that position-sensitive models perform better than position-insensitive models on average, indicating the effectiveness of using relative position information. However, our method MGSAG hasn't utilized relative position information, aiming to alleviate the position bias problem in ECPE. In spite of this, MGSAG still outperforms the existing state-of-the-art methods. Especially, MGSAG achieves the best $F_1$ on the main task: emotion-cause pair extraction. The $F_1$ score of MGSAG on ECPE is \textbf{1.06\%} higher than that of ECPE-MLL, which indicates the efficiency of capturing multi-granularity semantic relations between clauses. 


For the two sub-tasks, MGSAG outperforms other baselines in terms of cause extraction compared with emotion extraction. This indicates that the effective clause representation learning based on MGSAG is beneficial to extract cause clauses and further facilitate the extraction of emotion-cause pairs.

\subsubsection{Results on $Test_{Bias}$ and $Test_{NoBias}$}

To evaluate if MGSAG is vulnerable when the causes are not in proximity to the emotion, we evaluate it on the two subsets as shown in 5.1.2. Table~\ref{tab:Test1_Test2_Results} shows the results on $Test_{Bias}$ and $Test_{NoBias}$. Noted that when we get the best results on the original test set as shown in Table~\ref{tab:All_Test_Results}, we use the same parameters to evaluate models on the two subsets ($Test_{Bias}$ and $Test_{NoBias}$). 

From Table~\ref{tab:Test1_Test2_Results} we observe that there is a significant gap (34$\sim$41\%) between the results on $Test_{Bias}$ and $Test_{NoBias}$, for all of the methods. One of the reasons should be the imbalanced data of $Test_{Bias}$ and $Test_{NoBias}$, which means the proportion of position-insensitive data is very small. More importantly, most of the methods exploit the relative position information explicitly or implicitly, leading to poor performance on $Test_{NoBias}$.

However, MGSAG outperforms existing state-of-the-art baselines on both of the two subsets ($Test_{Bias}$ and $Test_{NoBias}$), proving its generalization ability towards position-sensitive and position-insensitive data. Specially, the $F_1$ score of MGSAG on $Test_{NoBias}$ is \textbf{3.13\%} higher than that of ECPE-MLL. The results verify the effectiveness of capturing causal relations between clauses via multi-granularity semantics encoding.



\begin{figure*}[!htbp]
	\centering
	\includegraphics[scale=0.33]{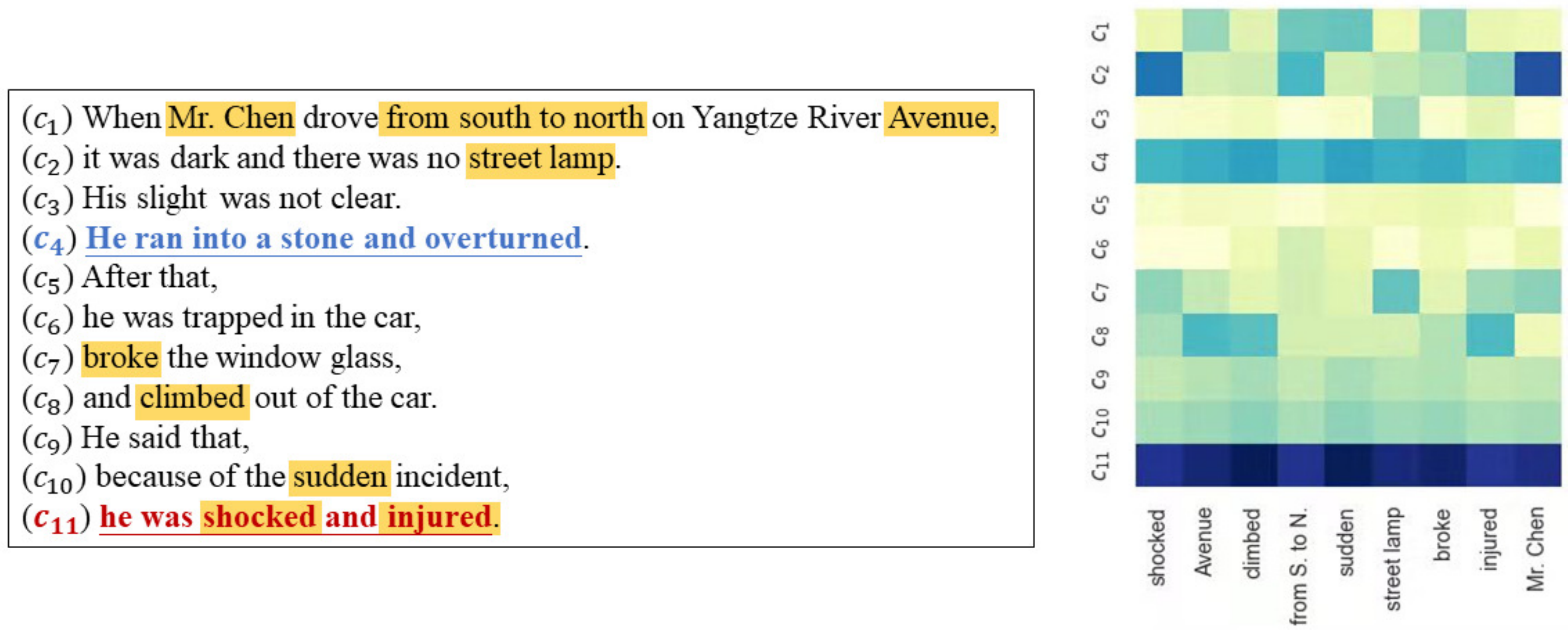}
	\caption{An example that MGSAG extracts the emotion cause pair $(c_{11}, c_4)$ correctly, while ECPE-MLL fails. Words shaded in yellow are keywords. The heatmap presents attention scores in the clause-keyword bipartite graph. Rows of $c_{11}$ and $c_4$ are the top-two darkest rows, means that keywords pay more attention to them and facilitate MGSAG to extract pair $(c_{11}, c_4)$ correctly.}
	\label{fig:case_study}
\end{figure*}

\subsection{Discussions}

We conduct ablation studies to analyze the effects of different components and settings in our method MGSAG.


\subsubsection{Influence of Different Components}

As shown in Table~\ref{tab:Ablation_Study}, we remove FGSAG, CGSAG, and both of them respectively to verify the effectiveness of the proposed two graphs with the semantics of different granularity.

\noindent \textbf{Effect of Fine-Grained Semantic Aware Graph.} We remove the FGSAG to verify the effect of fine-grained semantic enhanced relations. Table~\ref{tab:Ablation_Study} shows that removing FGSAG results in significant performance degradation, indicating that it is indeed useful for pair prediction. Especially, the result of $F_1$ on $Test_{NoBias}$ decreases \textbf{4.07\%} without the FGSAG, proving its efficiency of alleviating position bias.




\noindent \textbf{Effect of Coarse-Grained Semantic Aware Graph.} We remove CGSAG which is used for coarse-grained semantic enhanced relations to verify its effect. Table~\ref{tab:Ablation_Study} reports that model without CGSAG results in a clear drop (\textbf{2.74\%/3.17\%}) on $Test_{NoBias}$ and $Test_{all}$, but a limited drop (\textbf{0.76\%}) on $Test_{Bias}$. It shows that modeling the coarse-grained semantic relations between clauses can alleviate position bias as well. 

\noindent \textbf{Effect of Semantic Aware Graph Model.} We further evaluate the effect of dual graph-based modules by removing FGSAG and CGSAG simultaneously. As shown in Table~\ref{tab:Ablation_Study}, the model without the two graphs performs worse than without any one of them. The significant performance decline of the $F_1$ score on all of the test sets verifies that the fine-grained semantics and coarse-grained semantics are complementary to each other. Thus, it's necessary to take both of them into account.

\begin{table}\small
    \centering
    \renewcommand\arraystretch{1.0}
    \setlength{\tabcolsep}{7pt}{
    \begin{tabular}{c|c|c|c}
    \toprule
        Loss Function & P & R & $F_1$ \\
         
         
        \midrule[0.7pt]
        $\mathcal{L}_{pair}$ & 0.6940 & \textbf{0.6533} & 0.6720 \\
        \midrule[0.3pt]
        $\mathcal{L}_{pair} + \mathcal{L}_{emo} + \mathcal{L}_{cau}$ & \textbf{0.7243} & 0.6507 & \textbf{0.6846} \\
        
    \bottomrule
    \end{tabular}
    }
    \caption{Comparison of different supervised signals for our method.}
    \label{tab:Ablation_Supervision}
\end{table}

\subsubsection{Influence of Two-Level Supervision} We use the two-level supervised signals to train MGSAG. A low-level signal $\mathcal{L}_{emo} + \mathcal{L}_{cau}$ supervises the clause representation learning at the clause-level encoder and a high-level signal $\mathcal{L}_{pair}$ supervises the pair representation learning at the classification stage. To evaluate the effectiveness of low-level supervision, we only use $\mathcal{L}_{pair}$ to train the model, and the results are shown in Table~\ref{tab:Ablation_Supervision}. It shows that training with low-level supervision brings an improvement mainly on precision, which indicates that the low-level supervision is helpful to learn more accurate emotion-specific and cause-specific features and eventually facilitates the performance on emotion-cause pair extraction.

\begin{table}\small
    \centering
    \renewcommand\arraystretch{1.0}
    \setlength{\tabcolsep}{8pt}{
    \begin{tabular}{c|c|c|c}
    \toprule
        Model & $Test_{Bias}$ & $Test_{NoBias}$ & $Test_{all}$ \\
         
         
        \midrule[0.7pt]
        w/ RW & 0.7596 & 0.4078 & 0.6674 \\
        \midrule[0.3pt]
        w/o EW & 0.7669 & 0.3920 & 0.6686 \\
        \midrule[0.3pt]
        
        w/o TW & 0.7658 & 0.4271 & 0.6771 \\
        \midrule[0.7pt]
        
        \textbf{MGSAG} & \textbf{0.7730} & \textbf{0.4301} & \textbf{0.6846} \\

    \bottomrule
    \end{tabular}
    }
    \caption{Comparative $F_1$ results on $Test_{Bias}$, $Test_{NoBias}$, and $Test_{all}$ of our variant models, focusing on EC Pair Ext. ``w/ RW'' means using random embeddings for keyword feature initialization. ``w/o EW'' and ``w/o TW'' means removing emotion words and key phrases obtained by TextRank, respectively.}
    \label{tab:keyword_ablation}
\end{table}

\subsubsection{Influence of Different Keyword Settings} As shown in Table~\ref{tab:keyword_ablation}, we use different keyword settings to verify the effectiveness of our proposed keywords, which is the union of emotion words obtained from a sentiment lexicon \cite{SentiDict:Xu08} and key phrases obtained by TextRank \cite{TextRank:Mihalcea04EMNLP}. Removing any one of them results in a performance decline on all of the test sets. It proves that it's necessary to take both of them into account. Moreover, we replace the keyword features with randomly initialized embeddings, showing a significant drop on $Test_{NoBias}$. It indicates that the fine-grained semantics implied in keywords does help to alleviate the position bias problem.

\subsubsection{Case Study}

As shown in Figure~\ref{fig:case_study}, the distance between the emotion clause $c_{11}$ and the cause clause $c_4$ is 7. Although the cause clause $c_4$ doesn't contain any keywords, global keywords in the document convey crucial fine-grained semantics, helping MGSAG extracts $(c_{11}, c_4)$ correctly. 




\section{Conclusion and Future Work}

In this paper, we propose MGSAG to alleviate the position bias problem in the ECPE task. Our approach implements clause representation learning via fine-grained semantics introduced by keywords and coarse-grained semantics among clauses. Experimental results show that MGSAG surpasses the state-of-the-art baselines, and outperforms other methods significantly on the position-insensitive data. In the future, we would like to tackle the problem of imbalanced data by reducing non-emotion-cause pairs, based on a position-insensitive approach.
\bibliography{anthology,custom}
\bibliographystyle{acl_natbib}

\appendix

\section{Experimental Results with BERT}

\begin{table}[!htbp]\small
    \centering
    \renewcommand\arraystretch{1.0}
    \setlength{\tabcolsep}{10pt}{
    \begin{tabular}{l|c|c}
    \toprule
          Model & $Test_{Bias}$ & $Test_{NoBias}$ \\
        \midrule[0.7pt]
        
        PairGCN &  0.7246 & 0.3355  \\
        
        \textbf{MGSAG} & \textbf{0.7730} &  \textbf{0.4301} \\
        
        \midrule[0.7pt]
        
        PairGCN (BERT) & \textbf{0.8219} & 0.4005 \\
        

        \textbf{MGSAG (BERT)} & 0.8214 & \textbf{0.5004} \\
        
    \bottomrule
    \end{tabular}
    }
    \caption{$F_1$ results of varying approaches with and without BERT on $Test_{Bias}$ and $Test_{NoBias}$, focusing on emotion cause pair extraction.}
    \label{tab:BERT_res_Test1_Test2}
    \vspace{-1\baselineskip}
\end{table}

We implement MGSAG with the pre-trained BERT \cite{BERT:Devlin19NAACL} to explore the effect of pre-trained language model, where we use the base Chinese model\footnote{We download the pre-trained model from this link: \url{https://s3.amazonaws.com/models.huggingface.co/bert/bert-base-chinese.tar.gz}}. We replace the word-level encoder with the [CLS] embeddings of a clause which is obtained by BERT. Results on $Test_{Bias}$ and $Test_{NoBias}$ with and without BERT are shown in Table~\ref{tab:BERT_res_Test1_Test2}. Results on the original test set with and without BERT are shown in Table~\ref{tab:BERT_res_original}. 

During the training process, we use the Adam \cite{Adam:Kingma14ICLR} optimizer to update all parameters. The mini-batch size with BERT is set to 2. The learning rate with BERT is set to 1e-5.

\begin{table}[!tbp]\small
    \centering
    \renewcommand\arraystretch{1.0}
    \setlength{\tabcolsep}{8pt}{
    \begin{tabular}{l|c|c|c}
    \toprule
        
        \multirow{2}{*}{Model} & \multicolumn{3}{|c}{\textbf{Emotion Ext.}} \\
        \cline{2-4}
            ~ & P & R & $F_1$ \\
            
         \midrule[0.7pt]
         
        ECPE-2D & 0.8512 & 0.8220 & 0.8358  \\
        PairGCN & 0.8587 & 0.7208 & 0.7829  \\
        
        
        RankCP & 0.8703 & 0.8406 & \textbf{0.8548} \\
        
        ECPE-MLL & 0.8582 & \textbf{0.8429} & 0.8500  \\
        
        \midrule[0.3pt]
         \textbf{MGSAG} & \textbf{0.8721} & 0.7911 & 0.8287  \\
         
        \midrule[0.7pt]

        ECPE-2D (BERT) & 0.8627 & \textbf{0.9221} & 0.8910 \\
        
        PairGCN (BERT) & 0.8857 & 0.7958 & 0.8375 \\
        
        
        RankCP (BERT) & 0.9123 & 0.8999 & \textbf{0.9054} \\
        
        ECPE-MLL (BERT) & 0.8608 & 0.9191 & 0.8886 \\
         
        \midrule[0.3pt]

        \textbf{MGSAG (BERT)} & \textbf{0.9208} & 0.8211 & 0.8717 \\ 
        
        \bottomrule
        \toprule
        
        \multirow{2}{*}{Model} & \multicolumn{3}{|c}{\textbf{Cause Ext.}} \\
        \cline{2-4}
            ~ & P & R & $F_1$ \\
         
         \midrule[0.7pt]
        
        ECPE-2D & 0.7272 & 0.6298 & 0.6738   \\
        PairGCN & 0.7283 & 0.5953 & 0.6541   \\
        
        
        RankCP & 0.6927 & \textbf{0.6743} & 0.6824 \\
        
        ECPE-MLL & 0.7248 & 0.6702 & 0.6950   \\
        
        \midrule[0.3pt]
         \textbf{MGSAG} & \textbf{0.7510} & 0.6713 & \textbf{0.7080}  \\
         
        \midrule[0.7pt]

        ECPE-2D (BERT) & 0.7336 & 0.6934 & 0.7123 \\
        
        PairGCN (BERT) & 0.7907 & 0.6928 & 0.7375 \\
        
        
        RankCP (BERT) & 0.7461 & 0.7788 & 0.7615 \\
        
        ECPE-MLL (BERT) & 0.7382 & \textbf{0.7912} & 0.7630 \\
         
        \midrule[0.3pt]

        \textbf{MGSAG (BERT)} & \textbf{0.7979} & 0.7468 & \textbf{0.7712} \\
        
        \bottomrule
        \toprule
         
        \multirow{2}{*}{Model} & \multicolumn{3}{|c}{\textbf{Emotion Cause Pair Ext.}} \\
        \cline{2-4}
            ~ & P & R & $F_1$ \\
         
         \midrule[0.7pt]

         

        ECPE-2D & 0.6960 & 0.6118 & 0.6496  \\
        PairGCN & 0.6999 & 0.5779 & 0.6321  \\
        
        
        RankCP & 0.6698 & \textbf{0.6546} & 0.6610 \\
        
        ECPE-MLL & 0.7090 & 0.6441 & 0.6740  \\
        
        \midrule[0.3pt]
         \textbf{MGSAG}  & \textbf{0.7243} & 0.6507 & \textbf{0.6846} \\
         
        \midrule[0.7pt]

        ECPE-2D (BERT) & 0.7292 & 0.6544 & 0.6889 \\
        
        PairGCN (BERT)  & 0.7692 & 0.6791 & 0.7202 \\
        
        
        RankCP (BERT) & 0.7119 & \textbf{0.7630} & 0.7360 \\
        
        ECPE-MLL (BERT) & 0.7700 & 0.7235 & 0.7452 \\
         
        \midrule[0.3pt]

        \textbf{MGSAG (BERT)}  & \textbf{0.7743} & 0.7321 & \textbf{0.7521} \\
        
    \bottomrule
    \end{tabular}
    }
    \caption{Comparison of varying approaches with and without BERT on the original test set ($Test_{all}$).}
    \label{tab:BERT_res_original}
    \vspace{-1\baselineskip}
\end{table}

As shown in Table~\ref{tab:BERT_res_original}, methods with BERT perform better than those without BERT on the original test set, which shows the effectiveness of utilizing the pre-trained BERT. As shown in Table~\ref{tab:BERT_res_Test1_Test2}, results of models with BERT on $Test_{Bias}$ and $Test_{NoBias}$ indicate that using BERT as the encoder cannot make up for the deficiency caused by position bias. MGSAG still outperforms other methods on $Test_{all}$ and $Test_{NoBias}$. The results verify the effectiveness of capturing the causal semantic relations between clauses via fine-grained and coarse-grained semantics encoding.

\end{document}